\documentclass{new_tlp}
\usepackage{mathptmx}
\usepackage{amsmath}
\usepackage{algorithm,algpseudocode}
\usepackage{graphicx}

\usepackage{fancyvrb}
\usepackage{times}
\usepackage{soul}
\usepackage{url}
\usepackage{booktabs}
\usepackage{subcaption} 
\newtheorem{exmp}{Example}
\algnewcommand\algorithmicinput{\textbf{Input:}}
\algnewcommand\Input{\item[\algorithmicinput]}
\algnewcommand\algorithmicoutput{\textbf{Output:}}
\algnewcommand\Output{\item[\algorithmicoutput]}
\algnewcommand\algorithmicforeach{\textbf{for each}}
\algdef{S}[FOR]{ForEach}[1]{\algorithmicforeach\ #1\ \algorithmicdo}

\DeclareMathOperator*{\argmax}{argmax}

\algdef{S}[FOR]{ForEach}[1]{\algorithmicforeach\ #1\ \algorithmicdo}

\begin{document}
  \title[White-box Induction From SVM Models]
        {White-box Induction From SVM Models: Explainable AI with Logic Programming  }

  \author[F. Shakerin and G. Gupta]
         {FARHAD SHAKERIN, GOPAL GUPTA\\
         The University of Texas at Dallas, Texas, USA\\
         \email{$\{$Farhad.Shakerin, Gopal.Gupta$\}$@utdallas.edu}}

\newtheorem{lemma}{Lemma}[section]

\nocite{*}

\maketitle


%
\title{White-box Induction From SVM Models}

%

%

%
\begin{abstract}
We focus on the problem of inducing logic programs that explain models learned by the support vector machine (SVM) algorithm. The top-down sequential covering inductive logic programming (ILP) algorithms (e.g., FOIL) apply hill climbing search using heuristics from information theory. A major issue with this class of algorithms is getting stuck in a local optima. In our new approach however, the data dependent hill-climbing search is replaced with a model-dependent search where a globally optimal SVM model is trained first, then the algorithm looks into support vectors as the most influential data points in the model, and induces a clause that would cover the support vector and points that are most similar to that support vector. Instead of defining a fixed hypothesis search space, our algorithm makes use of SHAP, an example-specific interpreter in explainable AI, to determine relevant set of features. This approach yields an algorithm that captures SVM model's underlying logic and outperforms 
other ILP algorithms in terms of the number of induced clauses and classification evaluation metrics. This paper is under consideration for publication in the journal of ``Theory and practice of logic programming''.
\end{abstract}

%
%

%
\keywords{Explainable AI, Data Mining, Inductive Logic Programming, Machine Learning}

%

%

\section{Introduction}
Dramatic success of machine learning has led to an avalanche of applications of Artificial Intelligence (AI). However, the effectiveness of these systems is limited by the machines' current inability to explain their decisions and actions to human users. That is mainly because the statistical machine learning methods produce models that are complex algebraic solutions to optimization problems such as risk minimization or geometric margin maximization. Lack of intuitive descriptions makes it hard for users to understand and verify the underlying rules that govern the model. Also, these methods cannot produce a justification for a prediction they arrive at for a new data sample. 
The problem of explaining (or justifying) a model's decision to its human user is referred to as the model interpretability problem. The sub-field is referred to as Explainable AI.

The ILP learning problem is the problem of searching for a set of logic programming clauses that deduce the training examples. 
ILP is a thriving field and a large number of such clause search algorithms have been devised \cite{ilp20}. The search 
in these ILP algorithms 
is performed either top down or bottom-up. A bottom-up approach builds most-specific clauses from the training examples and searches the hypothesis space by using generalization. This approach is not applicable to large-scale datasets, nor it can incorporate \textit{negation-as-failure} \cite{Baral} into the hypotheses. A survey of bottom-up ILP systems and their shortcomings can be found at \cite{sakama05}. In contrast, top-down approach starts with the most general clauses and then specializes them. A top-down algorithm guided by heuristics is better suited for large-scale and/or noisy datasets \cite{quickfoil}.

\begin{figure}
    \includegraphics[width=0.9\textwidth,scale=0.2]{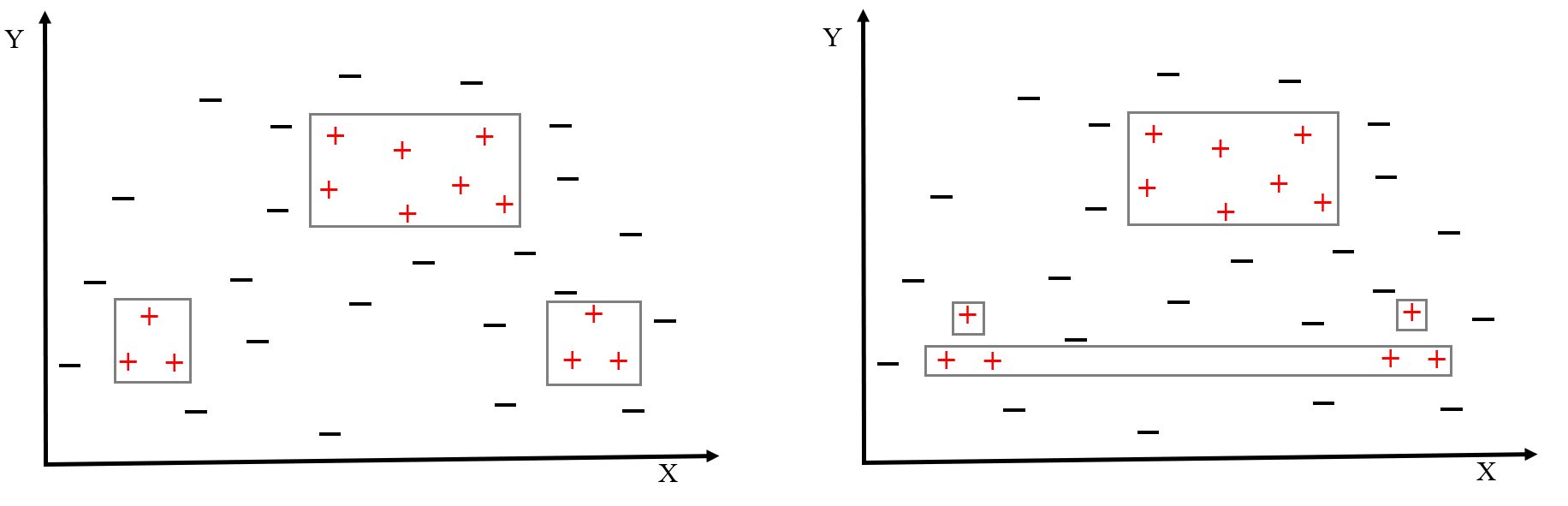}
\caption{Optimal sequential covering with 3 Clauses (Left), Sub-Optimal sequential covering with 4 Clauses (Right)}
\label{fig:localoptima}
\end{figure}

The FOIL algorithm by Quinlan \cite{foil} is a popular top-down algorithm. FOIL uses  heuristics from information theory called \textit{weighted information gain}. The use of a greedy heuristic allows FOIL to run much faster than bottom-up approaches and scale up much better. However, scalability comes at the expense of losing accuracy if the algorithm is stuck in a local optima and/or when the number of examples is insufficient. The former is an inherent problem in hill climbing search and the latter is due to the shrinking of examples during clause specialization. Figure \ref{fig:localoptima} demonstrates how the local optima results in discovering sub-optimal rules that do
not necessarily coincide with the real sub-concepts they are supposed to capture. 

Unlike top-down ILP algorithms, Support Vector Machine (SVM) \cite{svm} is a globally optimal learning method that generalizes very well and comes with 
test error upper-bound in terms of the number of support vectors and size of training input. However, this unique property is overshadowed by the black-box nature of SVM models. Explaining the behavior of black-box models has motivated a long line of research in Rule Induction from SVM models. As we argue in more detail in section \ref{sec:related}, all proposed Rule Extraction techniques either treat the model as black-box \cite{minerva,iter}, or are limited to certain type of kernels \cite{fung}, or are too complex to interpret \cite{svmprototype}. A survey of existing Rule Extraction techniques can be found in \cite{svmruleextraction}. 

Our new approach is based on the idea that each data sample is measurably similar/dissimilar to each support vector. Therefore, each support vector represents a subset of data samples. Now, if a set of features discriminate a support vector well, they would discriminate data samples similar to that support vector too. In order to measure the similarity and to pick up the support vector that is most similar to each data sample, we define a quantity based on the kernel value and each support vector's $\alpha$ parameter. To discover the most relevant features, our algorithm incorporates the SHAP technique. SHAP \cite{shap} is an example specific model interpreter that takes a model and an individual example and returns the contribution of each feature value in model's classification decision. 

This paper makes the following novel contribution: it introduces a novel ILP algorithm called SHAP-FOIL that iteratively learns a single clause for the most influential support vector (i.e., the support vector that pulls the highest number of similar data samples), then removes the covered examples including the support vector itself, and repeats this process until all examples/support vectors are covered. 
Our main claim is that our novel algorithm (SHAP-FOIL) learns a logic program that captures the underlying logic of an SVM model. This logic program serves as a human-understandable explanation of the SVM model. Our experiments using UCI datasets support our claim, as they show that the logic program that SHAP-FOIL learns is human understandable yet has precision, recall, accuracy and F$_1$-score that is comparable to the original SVM model. Explaining the model as a logic program leads to greater comprehensibility by humans as logic programming has a well-defined declarative and operational semantics. This is in contrast to other rule extraction methods that extract if-then-else rules that tend to have \textit{ad hoc} semantics.

Our work also shows how ILP can leverage the power of statistical machine learning methods to perform the search for induced logic programs significantly better. Our statistical machine learning based ILP method, SHAP-FOIL, performs significantly better than state-of-the-art ILP systems such as ALEPH \cite{aleph}. This claim, again, is supported by our experiments done using the UCI datasets, as they  show that the logic programs that are learned by SHAP-FOIL have precision, recall, accuracy, and F$_1$-score that are significantly better than ALEPH. In addition, a logic program that SHAP-FOIL learns is  more concise than the program learned by the ALEPH system for the same training set. Thus, SHAP-FOIL also improves comprehensibility (as measured by program conciseness, i.e., program size).

Rest of the paper is organized as follows. Section \ref{sec:background} gives  background on various concepts: explainability, ILP, the FOIL algorithm,  SVM, and the game theory-based SHAP method. Section \ref{sec:Shapfoil} presents our novel SHAP-FOIL algorithm. Section \ref{sec:Experiments} presents our experimental results, while Sections \ref{sec:related} discusses related work. Finally, Section \ref{sec:concl} presents our conclusions and planned future work.

\section{Background}
\label{sec:background}

\subsection{Explainable AI}

Statistical machine learning techniques learn a model from the training data with high precision and accuracy. Once a model is learned, a new, previously unseen input can be given to the model and result obtained. However, these models turn out to be ``black boxes" whose internal workings are not understood since they provide no clarity as to how the computed results depend on the input data. 

Consider an example where one wants to build a machine learning system to automate the process of approving consumer loans. Training data based on past customers (their income, age, years in job, repayment history, etc.) and the outcome for each (loan approved or declined) will be collected and a machine learning algorithm applied to it. The machine learning algorithm will learn a model to which a new, previously unseen customer's data can be input to determine if this new customer's loan application ought to be approved or not. The model will compute an output that tells us if a customer's loan application should be approved or declined, but it won't tell us or explain to us why. Being able to explain the model's decision is a major problem and its study is referred to as Explainable AI. 

We want explainability of the results computed by the model for many reasons some of which we list here: 

\begin{itemize}
    \item 
It allows users to understand why a particular result was produced and not something else.
\item If the result is justifiable, the user will trust the result more. 
\item It allows users to abide by the law as in many countries machine generated decisions have to come with a justification (e.g, European Union's GDPR).
\item It allows a user to judge if the results are correct since if the training data is incorrect or biased, results produced by the model will also be incorrect or biased.
\end{itemize}

Aside from explainability, we want to ensure that the rules are \textit{comprehensible} by humans, show \textit{fidelity} and are \textit{accurate} \cite{Martens2008}:
\begin{itemize} 
\item \textbf{Comprehensibility}: The extent to which extracted representations are
understandable by humans.
\item \textbf{Fidelity}: The extent to which the extraction representations model the black box.
\item \textbf{Accuracy}: The ability of extracted representations to make accurate
predictions on previously unseen cases.
\end{itemize}

Comprehensibility is typically measured by the number of rules produced. The assumption is that the fewer the rules, the easier it is for humans to understand the set of rules generated. One can also argue that logic programs are more comprehensible than other formalisms for expressing rules. This is because logic programming has a very well defined declarative and operational semantics. 

Note that precision, recall and F$_1$ score are the typical metrics used to measure fidelity. 


\subsection{Inductive Logic Programming}
Inductive Logic Programming (ILP) \cite{ilp} is a subfield of machine learning that learns models in the form of logic programming rules (Horn Clauses) that are comprehensible to humans. This problem is formally defined as:\\
\textbf{Given}
\begin{enumerate}
    \item a background knowledge theory $B$, in the form of an extended logic program, i.e., clauses of the following form:\\
    $~~~~~~~~~~~~h \leftarrow l_1, ... , l_m,\ \textbf{not} \ l_{m+1},...,\ \textbf{not} \ l_n$ \\
    where $l_1,...,l_n$ are positive literals and \textbf{not} denotes \textit{negation-as-failure} (NAF) \cite{Baral} and $B$ has no \textit{even cycle} \cite{gelfondkahl}
    \item two disjoint sets of ground target predicates $E^+, E^-$ known as positive and negative examples, respectively
    \item a hypothesis language of function free predicates $L$, and a  refinement operator $\rho$ under $\theta$-subsumption \cite{plotkin70} that would disallow even cycles.
\end{enumerate}
\textbf{Find} a set of clauses $H$ such that:
\begin{itemize}
    \item $ \forall e \in \ E^+ ,\  B \cup H \models e$
    \item $ \forall e \in \ E^- ,\  B \cup H \not \models e$
    \item $B \land H$ is consistent.
\end{itemize}

While we allow normal logic programs in our definition, we only learn logic programs with no negated literals in our current work (any negated calls in these learned logic program arise due to propositionalisation or one-hot encoding and can be executed using negation as failure as supported in Prolog). Even cycle refers to programs where a recursive call occurs in the scope of even number of negation, for example, the following program has an even cycle.

\begin{verbatim}
    p :- not q.
    q :- not p.
\end{verbatim}

$\theta$-subsumption \cite{plotkin70} is defined as follows: A clause $c$ theta-subsumes another clause $d$ if and only if there exists a variable substitution $\theta$ such that $c \theta \subseteq d$. $\theta$-subsumption determines a partial order of the clause space. Top element of this partial order is the most general clause $p(X_1,...,X_n) \gets true.$. A refinement operator $\rho$ which moves down this partial order, takes a hypothesis clause $c$ and gives back a set of all syntactically modified version $c'$ of $c$. In other words, each clause $c'$ is a $\theta$-subsumed specialization of $c$. For instance, the refinement operator $\rho$ in Algorithm \ref{algo:foil} specializes the clause $h \gets b_1,...b_n .$ by adding a new literal $l$ to the clause yielding $h \gets b_1,...b_n,l$.  




A large number of ILP algorithms have been devised \cite{ilp20}. FOIL \cite{foil} is one such illustrative algorithm for ILP that is well known. 
FOIL is a top-down ILP algorithm which follows a \textit{sequential covering} scheme to induce a hypothesis. The FOIL algorithm is summarized in Algorithm \ref{algo:foil}. This algorithm repeatedly searches for clauses that score best with respect to a subset of 
positive and negative examples, a current hypothesis and a heuristic called \textit{information gain} (IG).

\begin{algorithm}
\caption{Summarizing the FOIL algorithm}
\label{algo:foil}
\begin{algorithmic}[1]
\Input $target,B,E^+,E^-$ 
\Output 
Initialize $H \gets \emptyset $
\While{($|E^+| > 0$)}
	\State $c \gets (target$ :- $ \ true.)$
	\While{($|E^-| > 0 \land c.length < max\_length $)}
		\For{all $ \ c' \in \rho (c)$}
        	\State $compute \ score(E^+,E^-,H \cup \{c'\},B)$
    	\EndFor
    	\State let $\hat{c}$ be the $c' \in \rho(c)$ with the best score   
         \State $E^- \gets covers(\hat{c},E^-)$
    \EndWhile	
    \State add $\hat{c}$ to $H$
    \State $E^+ \gets E^+ \setminus covers(\hat{c},E^+)$
\EndWhile 
\State \textbf{return} $H$
\end{algorithmic}
\end{algorithm}

Starting from the most general clause (i.e., $p(X_1,...,X_n) \gets true.$ where the predicate $p/n$ is the predicate being learned and each $X_i$ is a variable), the inner loop searches for a clause with the highest information gain using a general-to-specific hill-climbing search. To find the ``best" clause, first, a refinement operator $\rho$ specializes the current clause $h \gets b_1,...b_n .$ by adding a new literal $l$ to the clause yielding $h \gets b_1,...b_n,l$. The ``best" choice for literal $l$ is decided by a heuristic based search that employs information gain. 
In FOIL, information gain for a given clause is calculated as follows: 
\begin{equation}
IG(L,R) = t\left(log_2 \frac{p_1}{p_1 + n_1} - log_2 \frac{p_0}{p_0+ n_0} 
\right)
\end{equation}
where $L$ is the candidate literal to add to rule $R$, $p_0$ ($n_0$) is the number of 
positive (negative) examples covered by $R$ respectively, $p_1$ ($n_1$) is the number of positive (negative) examples covered by $R+L$ respectively, and $t$ is the number of positive examples that are covered by $R$ and $R+L$ together. The function $covers$ in lines 8 and 11, takes a clause $c$ and a set of examples $E$, and returns another (possibly empty) set of examples that are logically implied by $c$. 

\subsection{Support Vector Machines}
Given a training dataset of $m$ data points 
$\vec{x}_i \in {\rm I\!R}^n$ and $m$ corresponding labels $y_i \in \{1,-1 \}$, the linear support vector machine is defined as the following optimization problem:

\noindent
\begin{equation}
\label{eq:primal}
\min_{\vec{w},b,\xi_i \ge 0} \frac{1}{2} \| \vec{w}\|_2 ^2 + C \sum_{i=1}^{m} \xi_i    
\end{equation}
such that: $ y_i(\vec{w} . \vec{x}_i - b) + \xi_i \ge 1$
where $\xi_i$ is the slack error to potentially allow some points to be misclassified, $\vec{w}$ is the perpendicular vector to the separating hyper-plane, $b$ is the offset of that hyper-plane, $C$ is a hyperparameter and determines the degree to which misclassification is allowed to avoid over-fitting. 

An equivalent yet more efficient form of SVM problem known as dual formulation is defined as follows:
\begin{equation}
 \label{eq:dual}
 \max_{\alpha_i} -\frac{1}{2}\sum_{i=1}^{m}y_i y_j\vec{x}_i \vec{x}_j \alpha_i \alpha_j + \sum_{i=1}^{m} \alpha_i  
\end{equation}
such that $\sum_{i=1}^{m} \alpha_i y_i = 0$ and $0 \le \alpha_i \le C$

\noindent
The dot product in dual formulation can be replaced with any kernel function (i.e., a function that maps data into higher dimensions). For non-linearly
separable data, kernels map the data into a higher dimensional feature space where data becomes separable. Equation \eqref{eq:dual} is a special case for the following general dual formulation:
\begin{equation}
\label{eq:dualkernel}
\max_{\alpha_i} -\frac{1}{2}\sum_{i=1}^{m}y_i y_j K(\vec{x}_i, \vec{x}_j) \alpha_i \alpha_j + \sum_{i=1}^{m} \alpha_i      
\end{equation} 
such that $\sum_{i=1}^{m} \alpha_i y_i = 0$ and $0 \le \alpha_i \le C$

\noindent
After solving the above problem using quadratic programming, the $\alpha_i$ and $b$ are used to classify a new data sample $\vec{x}$ as follows:
\begin{equation}
\label{eq:classify}
f(x) = sign \bigg [\sum_{i=1}^{m}  \alpha_i y_i K(\vec{x}_i, \vec{x}) + b \bigg ]    
\end{equation} 

It turns out that $\alpha$ is a sparse vector and only few $\alpha_i$ come back with non-zero values from the quadratic solver package. They are called the support vectors and as Equation \eqref{eq:classify} suggests, proportionate to their respective $\alpha_i$ value, support vectors are the only influential data points in classification decision of any new data sample.

\begin{figure}
    \includegraphics[width=0.45\textwidth]{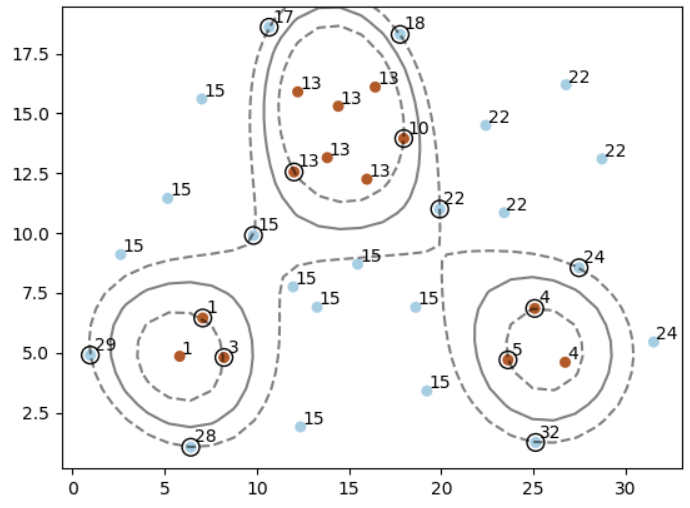}
\caption{Annotating Data Points in a 2D dataset With Most Similar Support Vector}
\label{fig:2d_similarity}
\end{figure}

While the kernel function is meant to map data points into a higher dimension efficiently, one can also interpret the kernel value $K(\vec{x}_i, \vec{x}_j)$ as a similarity measure between points $\vec{x}_i$ and $\vec{x}_j$. For instance, in case of the Gaussian radial basis kernel (rbf):
\begin{equation}
\label{eq:rbf}
K(\vec{x}_i, \vec{x}_j) = e^{-\gamma  \| \vec{x}_i - \vec{x}_j\|_2 ^2 }
\end{equation}
where $\gamma$ is a hyper parameter. If $\vec{x}_i$ and $\vec{x}_j$ are similar, the kernel value would be close to 1. Otherwise, it would be close to 0. This can be naturally used to quantify similarity. However, it should be noted from Equation \eqref{eq:classify} that the magnitude of $\alpha_i$ also contributes to the influence and similarity. Therefore, we define the similarity of data point $\vec{x}$ and the $i^{th}$ support vector $\vec{x}_i$ as follows:
\begin{equation}
\label{eq:similarity}
    sim_i(\vec{x}) = \alpha_i y_i K(\vec{x}_i, \vec{x})
\end{equation}
For any new data sample $\vec{x}$ the support vector with highest $sim$ value is the one that contributes most to the prediction of $\vec{x}$. Figure \ref{fig:2d_similarity} demonstrates an SVM model with rbf kernel trained on the same dataset from Figure \ref{fig:localoptima}. In this figure, support vectors are the dots located on the dashed lines. They are labeled with an integer identifier. Every other data point is annotated with the identifier of the most similar support vector calculated using Equation \eqref{eq:similarity}. Since the similarity is measured with respect to the concept being learned, the most similar support vector does not necessarily accord with the Euclidean distance.
\subsection{SHAP}
SHAP \cite{shap} (SHapley Additive exPlanations) is a unified approach with foundations in game theory to explain the output of any machine learning model in terms of its features' contributions. To compute each feature $i$'s contribution, SHAP requires retraining the model on all feature subsets $S \subseteq F$, where $F$ is the set of all features. For any feature $i$, a model $f_{S \cup \{i\}}$ is trained with the feature $i$ present, and another model $f_S$ is trained with feature $i$ eliminated. Then, the difference between predictions is computed as follows: $f_{S \cup \{i\}}(x_{S \cup \{i\}}) - f_S(x_S)$, where $x_S$ represents sample's feature values in $S$. Since the effect of withholding a feature depends on other features in the model, the above differences are computed for all possible subsets of $S \subseteq F \setminus \{i\}$ and their average taken. The weighted average of all possible differences (a.k.a Shapley value) is used as feature importance. Equation \eqref{eq:shapleynumber} shows how Shapley value associated with each feature value is computed:
\begin{equation}
\label{eq:shapleynumber}
\phi_i = \sum_{S \subseteq F \setminus \{i\}} \frac{|S|!(|F| - |S| -1)!}{|F|!} \bigg [f_{S \cup \{i\}}(x_{S \cup \{i\}}) - f_S(x_S) \bigg ]
\end{equation}

\begin{figure}[h]
    \includegraphics[width=0.95\textwidth]{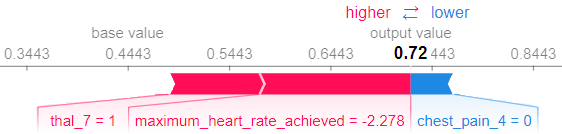}
\caption{Shap Values for A UCI Heart Prediction }
\label{fig:heartshap}
\end{figure}

 Given a dataset and a trained model, SHAP outputs a matrix with the shape $(\# samples, \#features)$ representing the Shapley value of each feature for each data sample. Each row sums to the difference between the model output for that sample and the expected value of the model output. This difference explains why the model is inclined on predicting a specific class outcome.
\begin{exmp}
\label{ex:heart}
The UCI heart dataset contains features such as patient's  blood pressure, chest pain, thallium test results, number of major vessels blocked, etc. The classification task is to predict whether the subject suffers from heart disease or not. Figure \ref{fig:heartshap} shows how SHAP would explain a model's prediction over a data sample.   
\end{exmp}
For this individual, SHAP explains why the model predicts heart disease by returning the top features along with their Shapley values (importance weight). According to SHAP, the model predicts ``heart disease" because of the values of ``thalium test" and ``maximum heart rate achieved" which push the prediction from the base (expected) value of 0.44 towards a positive prediction (heart disease). On the other hand, the feature ``chest pain" would have pushed the prediction towards negative (healthy), but it is not strong enough to turn the prediction. 

The categorical features should be \textit{binarized} before an SVM model can be trained. Binarization (aka one-hot encoding) is the process of transforming each categorical feature with domain of cardinality $n$, into $n$ new binary predicates (features). In Example \ref{ex:heart}, chest pain level is a categorical feature with 4 different values in the set $\{1,2,3,4\}$. Type 4 chest pain indicates asymptomatic pain and is a serious indication of a heart condition. In this case, binarization results in 4 different predicates. The ``thalium test" is also a categorical feature with outcomes in the set \{3,6,7\}. Any outcome other than 3, indicates a defect (6 for fixed and reversible for 7).  

In case of Example \ref{ex:heart}, SHAP determines that the feature ``thal\_7" with outcome of 1 (True), pushes the prediction towards heart disease. To reflect this fact in our SHAP-FOIL ILP algorithm (described next), for any person $X$, the predicate \texttt{thal(X,7)} is introduced. Also, SHAP indicates that the binary feature ``chest\_pain\_4" with value 0 (False), pushes the prediction towards healthy. In our SHAP-FOIL algorithm this is represented by \textit{negation-as-failure} \cite{Baral} as~ \texttt{not chest\_pain(X,4)}. 
Note that, at present, negation-as-failure is used only to succinctly represent such one-hot encoded predicates.

\section{The SHAP-FOIL Algorithm}
\label{sec:Shapfoil}

In this section we introduce SHAP-FOIL, an algorithm capable of learning 
logic programs based on the global behavior of an SVM model. 

There are two major issues with the \textit{sequential covering} algorithms such as FOIL: 1) As number of examples decreases during specialization loop, probability of introducing an irrelevant predicate that accidentally splits a particular set of examples increases.  2) The greedy nature of \textit{hill-climbing} search in clause specialization sometimes results in introduction of wrong predicates that would cover more examples at a certain moment, but eventually leads to inducing a clause that does not perfectly represent sub-concepts as shown in Example \ref{fig:localoptima}. This is known as Local Optima problem in \textit{hill-climbing} search. Determining the sub-concepts requires a global view which could only happen via a global optimization process such as an SVM model. However, finding the best separating hyperplane in a higher dimension does not explain the contributing features in any classification decision made by the model. 

SHAP is able to quantitatively explain the features that would push the model towards predicting a specific outcome. In particular, for each support vector, SHAP determines a subset of feature value pairs that would make the model arrive at a certain decision. It turns out that just by having the Shapley values of support vectors, our algorithm can learn the global underlying behavior of SVM model. This is mathematically justified as follows: From Equation \eqref{eq:classify}, every new data sample is interpreted in terms of similarity to support vectors. The internal points are not relevant (because their corresponding $\alpha_i$ parameter is 0). Among support vectors, only the ones that are closely ``similar" to the given point are relevant (because, for dissimilar support vectors the kernel value of Equation \eqref{eq:rbf} is close to 0).   

The intuition behind SHAP-FOIL algorithm is as follows: If a subset of feature-values explains the decision on a particular support vector, it  explains the decision on data points that are ``similar" to that support vector too. Similarity is measured using Equation \eqref{eq:similarity}. In the context of \textit{sequential covering} scheme, SHAP-FOIL would find the support vector that pulls the greatest number of data points in terms of Equation \eqref{eq:similarity}. Next, it would specialize a clause by introducing predicates that are determined by SHAP for that support vector. Then, the algorithm removes the data points that are covered by that rule. It also removes the support vector. This process is repeated  for the remaining data points. Since there are only finite number of support vectors, the algorithm is guaranteed to terminate.

Unlike most ILP algorithms, SHAP-FOIL does not require discretization of numerical features in advance. During the specialization of a clause, if a numeric feature happens to have the highest Shapley number for a support vector, a real arithmetic constraint is introduced by the algorithm. The end-points of this interval is determined by looking into the respective values of all data points that are most similar to that support vector. 

\begin{algorithm}[ht]
\caption{Summarizing the SHAP-FOIL algorithm}
\label{algo:shapfoil}
\begin{algorithmic}[1]
\Input $D = \{ (\vec{x}_1,y_1),...,(\vec{x}_m,y_m) \}$
\Input $ SHAP~ matrix~(m,\#features), \Theta $
\Input $S = \{ sv ~ | ~ sv ~ is ~ a ~ support ~ vector ~ and ~ sv ~ is ~ TP \}$
\Output $Hypothesis~ H = \{\}$
\Function{SHAP-FOIL}{$S,D$}
    \While{($ S ~ \neq ~ \emptyset$)}
        \State $sim\_map$ = \Call{annotate\_samples}{$S$,$D$}
        \State $sv = \argmax_{s \in S} len(sim\_map[s])$
        \State $c \gets (target$ :- $ \ true.)$
        \State $\hat{c} \gets$ Specialize $c$ using SHAP[sv]
        \If {$\hat{c}$'s accuracy on D $\geq \Theta$}
            \State $H \gets H \cup \{\hat{c}\}$
            \State $D \gets D - \{\vec{x}_i ~| \vec{x}_i ~ \in D ~ \land \hat{c} \models y_i  \}$
        \EndIf
        \State $ S \gets S - \{ sv \}$
    \EndWhile
\State \textbf{return} $H$
\EndFunction
\\
\Function{annotate\_samples}{$S,D$}
    \State Let $ sim\_map$ be a map of type $:S \mapsto List$   
    \ForEach {$sv \in S $}
		\State $ sim\_map[sv] = [~]$
	\EndFor
    \ForEach {$\vec{x}_i \in D $}
		\State $ sv = \argmax_{s \in S} sim(s,\vec{x_i}) $
		\State $sim\_map[sv].append(\vec{x}_i)$
	\EndFor
\State \textbf{return} $sim\_map$
\EndFunction

\end{algorithmic}
\end{algorithm}

Algorithm \ref{algo:shapfoil} summarizes the SHAP-FOIL algorithm. The algorithm inputs are as follows: 1) Set $D$ of cardinality $m$ representing $m$ training data points. 2) SHAP matrix of a trained SVM model on $D$. 3) A threshold $\Theta$ to determine minimum acceptable accuracy of each induced clause. 4) Set $SV$ that are True Positive (TP) support vectors (i.e., support vectors with label 1, also predicted 1 by the SVM model). The model outputs a hypothesis comprising a set of induced clauses. The While loop in line 2, iterates until all support vectors are considered. (termination condition). In line 3, by calling the function ANNOTATE\_SAMPLES, all remaining data-points are annotated with a support vector among the remaining support vectors. This support vector must have the highest similarity value to that data point. In line 4, the algorithm chooses $sv$, the support vector which pulled the greatest number of data points from annotation. This allows to discover the more inclusive rules first. In lines 5 and 6, similar to the FOIL algorithm, specialization from the most general clause (i.e., \texttt{target :- true.}) is conducted. To specialize a clause, predicates determined by the Shapley value of $sv$ are added to the body in the order of their Shapley value magnitude. To add numeric features, our algorithm creates an interval by finding the smallest and largest values in the list of data samples associated with $sv$. The specialized clause is named $\hat{c}$. In line 7, the accuracy of $\hat{c}$ is tested against a threshold $\Theta$. If it is higher than $\Theta$, $\hat{c}$ is added to the current hypothesis $H$ in line 8. In line 9, similar to FOIL, the set of data points covered by $\hat{c}$ are removed from $D$ (\textit{sequential covering}). If $\hat{c}$ achieves lower accuracy than $\Theta$, it is discarded. Regardless of the case, in line 11, $sv$ is removed from the set of support vectors. This serves two purposes: 1) It guarantees the termination. 2) More importantly, if in some iteration, a support vector pulls greatest number of similar points but it does not yield an above $\Theta$ accurate clause, to make progress possible, this support vector will be removed from consideration. We will clarify this more in Example \ref{ex:ex3}.

\begin{exmp}
\label{ex:ex2}
Figure \ref{fig:localoptima} illustrates the local optima issue of FOIL. In Figure \ref{fig:2d_similarity}, an SVM model is shown for the same dataset with two features f1 and f2 and two classes of red and blue. The SHAP-FOIL algorithm learns the following logic program on this dataset for the class red. 
While there are six support vectors, the set of red example data points were found most similar to only three support vectors. As a a result, only three rules were generated for ``red" concept, shown below. The `blue' concept can be learned similarly.
\end{exmp}
\begin{verbatim}
red(X):- f1(X,F1), 12.02 =< F1 <= 17.97,
         f2(X,F2), 12.25 =< F2 <= 16.1 .
red(X):- f1(X,F1), 5.82 =< F1 <= 8.22,
         f2(X,F2), 4.8 =< F2 <= 6.45 .
red(X):- f1(X,F1), 23.62 =< F1 <= 26.72,
         f2(X,F2), 4.6 =< F2 <= 6.85 .
\end{verbatim}

To handle numeric features, our algorithm creates an interval by finding the smallest and largest values in the list of data samples associated with a support vector $sv$. This approach sometimes results in too coarse-grained intervals that cover too many False Positives (FP) to tolerate. As explained earlier, to handle this case, SHAP-FOIL removes $sv$ and tries to break the region into smaller sub-regions. Each smaller region is then covered by other support vectors. We illustrate this with an example.

\begin{exmp}
\label{ex:ex3}
  
\begin{figure}
\centering
    \includegraphics[width=0.8\textwidth]{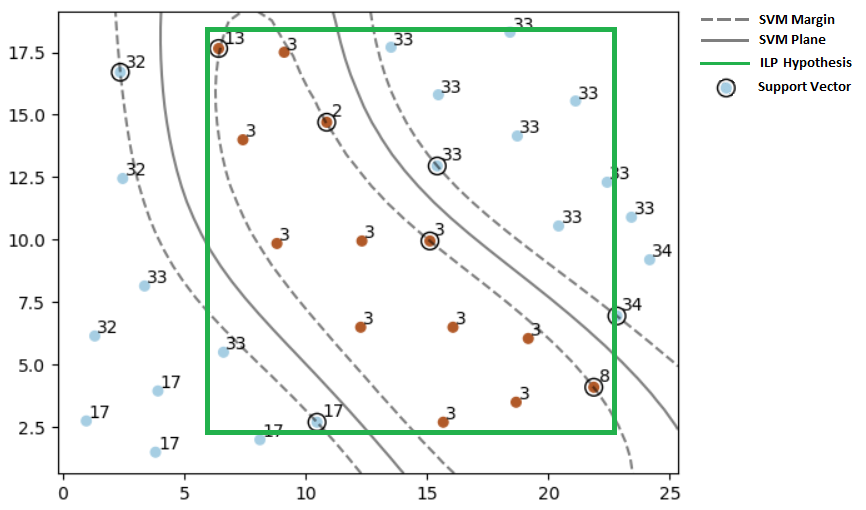}
\caption{Iteration \#1 of Example \ref{ex:ex3} }
\label{fig:ex3iter1}
\end{figure}
\begin{figure}
\centering
    \includegraphics[width=0.8\textwidth]{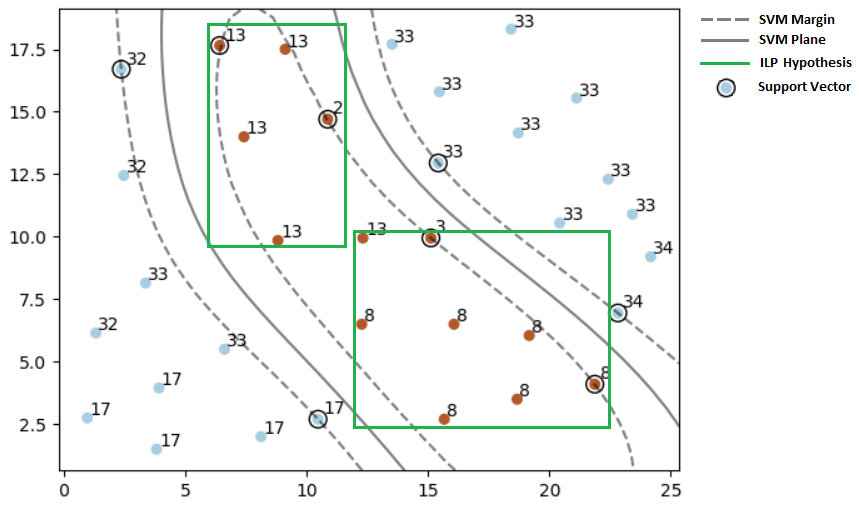}
\caption{Iteration \#2, \#3 of Example \ref{ex:ex3} }
\label{fig:ex3iter2}
\end{figure}

Consider a dataset of two features on which an SVM model is trained (Figure \ref{fig:ex3iter1}). At iteration 1, after annotating the data samples, the support vector 3, pulls the majority of data samples. According to SHAP, both numeric features contribute to the classification decision. However, after creating the intervals for both features, the clause shown as a green box, ends-up covering significant number of False Positives. This is shown in Figure \ref{fig:ex3iter1}. Therefore, the clause is discarded and the support vector 3, is removed from consideration. As shown in Figure \ref{fig:ex3iter2}, on iteration 2, the support vector 8 pulls the highest number of data samples. It yields an accurate clause. Thus, it is added to the hypothesis. On iteration 3, the support vector 2 pulls the rest of data samples and once again, it results in an accurate clause.
\end{exmp}

\section{Experimental Results}
\label{sec:Experiments}

In this section, we present our experiments on UCI standard benchmarks \cite{uci}. 
We compare the performance of the logic program output by SHAP-FOIL with that of the original SVM model and the ALEPH ILP system \cite{aleph}. 
ALEPH is a state-of-the-art ILP system that has been widely used in prior work. To find a rule, ALEPH starts by building the most specific clause, which is called the ``bottom clause", that entails a seed example. Then, it uses a branch-and-bound algorithm to perform a general-to-specific heuristic search for a subset of literals from the bottom clause to form a more general rule. We set ALEPH to use the heuristic enumeration strategy, and the maximum number of branch nodes to be explored in a branch-and-bound search to 500K. We use the standard metrics including precision, recall, accuracy and F$_1$ score to measure the quality of the results.

Table \ref{tbl:accuracies} presents the comparison between ALEPH and SHAP-FOIL on classification evaluation of each UCI dataset. It also shows the results for the SVM method. The best performer is highlighted with boldface font. With the exception of congressional voting dataset where the SVM performance is lower than ALEPH, the SHAP-FOIL algorithm always achieves higher score compared to ALEPH. 
%
Our experiments illustrate two important points:  

\begin{enumerate}
    \item A  logic program induced by SHAP-FOIL not only explains the black-box SVM model, it performs well wrt all classification evaluation metrics, thus exhibiting high fidelity (quantitatively measured by F$_1$-score).
    \item Our ILP algorithm in which learning of clauses is guided by a model-dependent search is superior in performance to traditional ILP algorithms such as ALEPH. This insight can be adopted by other ILP algorithms to make them better. 
\end{enumerate}

\begin{table}
{\scriptsize
\centering
\begin{tabular}{|l|c|c|c|c|c|c|c|c|c|c|c|c|c|}
\cline{1-14}
           & \multicolumn{1}{l|}{}       & \multicolumn{12}{c|}{Algorithm}                                                                                                                                                                                                                                                                                                                  \\ \cline{1-14}
           & \multicolumn{1}{l|}{}       & \multicolumn{4}{c|}{SVM}                                                                                       & \multicolumn{4}{c|}{Aleph}                                                                                     & \multicolumn{4}{c|}{SHAP-FOIL}                                                                                 \\ \cline{1-14}
Data Set   & \multicolumn{1}{l|}{Kernel} & \multicolumn{1}{l|}{Prec.} & \multicolumn{1}{l|}{Recall} & \multicolumn{1}{l|}{Acc.} & \multicolumn{1}{l|}{F$_1$} & \multicolumn{1}{l|}{Prec.} & \multicolumn{1}{l|}{Recall} & \multicolumn{1}{l|}{Acc.} & \multicolumn{1}{l|}{F$_1$} & \multicolumn{1}{l|}{Prec.} & \multicolumn{1}{l|}{Recall} & \multicolumn{1}{l|}{Acc.} & \multicolumn{1}{l|}{F$_1$} \\ \cline{1-14}
credit-j   & rbf                         & 0.84                       & 0.84                        & 0.84                      & 0.84                    & 0.78                       & 0.72                        & 0.78                      & 0.75                    & \textbf{0.83}              & \textbf{0.76}               & \textbf{0.83}             & \textbf{0.80}           \\ \cline{1-14}
breast-w   & Poly                        & 0.97                       & 0.96                        & 0.96                      & 0.96                    & 0.92                       & 0.87                        & 0.93                      & 0.89                    & \textbf{0.97}              & \textbf{0.89}               & \textbf{0.95}             & \textbf{0.93}           \\ \cline{1-14}
ecoli      & rbf                         & 0.96                       & 0.96                        & 0.96                      & 0.96                    & 0.85                       & 0.75                        & 0.84                      & 0.80                    & \textbf{0.86}              & \textbf{0.94}               & \textbf{0.89}             & \textbf{0.90}           \\ \cline{1-14}
kidney     & poly                        & 0.99                       & 0.99                        & 0.99                      & 0.99                    & 0.96                       & 0.92                        & 0.93                      & 0.94                    & \textbf{0.97}              & \textbf{0.97}               & \textbf{0.97}             & \textbf{0.97}           \\ \cline{1-14}
voting     & rbf                         & 0.95                       & 0.94                        & 0.94                      & 0.94                    & \textbf{0.97}              & \textbf{0.94}               & \textbf{0.95}             & \textbf{0.95}           & 0.92                       & 0.94                        & 0.91                      & 0.93                    \\ \cline{1-14}
autism     & rbf                         & 1.00                       & 1.00                        & 1.00                      & 1.00                    & 0.73                       & 0.43                        & 0.79                      & 0.53                    & \textbf{0.94}              & \textbf{0.86}               & \textbf{0.94}             & \textbf{0.88}           \\ \cline{1-14}
ionosphere & rbf                         & 0.95                       & 0.95                        & 0.94                      & 0.95                    & 0.89                       & 0.87                        & 0.85                      & 0.88                    & \textbf{0.92}              & \textbf{0.90}               & \textbf{0.90}             & \textbf{0.91}           \\ \cline{1-14}
heart      & poly                        & 0.81                       & 0.80                        & 0.80                      & 0.80                    & 0.76                       & 0.75                        & 0.78                      & 0.75                    & \textbf{0.90}              & \textbf{0.86}               & \textbf{0.90}             & \textbf{0.88}           \\ \cline{1-14}
\end{tabular}
\caption{Evaluation of SHAP-FOIL on UCI Datasets}
\label{tbl:accuracies}
}
\end{table}

The \textit{sequential-covering} based algorithms---including ALEPH and FOIL---tend to learn too many rules in presence of noisy data. Both algorithms induce more accurate clauses at the expense of covering fewer examples by each clause. In our SHAP-FOIL algorithm, specialization is stopped once the purity of a clause reaches the threshold $\Theta$ while the maximum coverage is guaranteed by SHAP because the specialization is performed in the order of feature's Shapley number. For instance, while ALEPH discovers 15 clauses for UCI heart, the following logic program comprised of only 6 clauses is induced by the SHAP-FOIL algorithm:
\begin{verbatim}
(1) heart_disease(X) :- thallium_test(X,7),
                        chest_pain(X,4),
                        exercise_induced_angina(X).
           
(2) heart_disease(X) :- maximum_heart_rate_achieved(X,F1),
                        106 =< F1, F1 =< 154,
                        not major_vessels(X,0),
                        oldpeak(X,F2),
                        1 =< F2, F2 =< 4.
(3) heart_disease(X) :- not major_vessels(X,0),
                        thallium_test(X,7),
                        chest_pain(X,4).
(4) heart_disease(X) :- thallium_test(X,7),
                        age(X,F1), 
                        35 =< F1, F1 =< 52,
                        chest_pain(X,4). 
(5) heart_disease(X) :- maximum_heart_rate_achieved(X,F1),
                        120 =< F1, F1 =< 147,
                        exercise_induced_angina(X),
                        chest_pain(X,4).
(6) heart_disease(X) :- not major_vessels(X,0),
                        chest_pain(X,4),
                        male(X).
\end{verbatim}  

The induced program can be understood as follows: In clause (1), \texttt{thallium\_test(X,7)} indicates a \textit{thallium test} with reversible defect, while \texttt{chest\_pain(X,4)} indicates an asymptomatic type of chest pain. According to clause (1), these two conditions, conjoined with angina revealed in an exercise test indicate the existence of heart disease. In clause (2), if maximum heart rate is achieved and during exercise test it falls in the discovered range 106-154, ST depression induced by exercise relative to rest falls in the range 1-4, and there are signs of blockage in major vessels (indicated by \textit{negation-as-failure}), then the combination means heart disease. The rest of the clauses are read similarly. 

As stated earlier, our SHAP-FOIL algorithm not only does better than ALEPH in classification evaluation measures, it also produces much smaller number of rules. In many cases, our SHAP-FOIL algorithm produces an order of magnitude fewer rules than the ALEPH system. Obviously, smaller number of rules are more readily understood by users. They can be manually revised by the user much more easily as well (to better capture the learned knowledge) based on user's background knowledge about the problem.
Thus, logic programs discovered by our SHAP-FOIL algorithm are more comprehensible as they are shorter. Table \ref{tab:progsize} shows the size of the program discovered by ALEPH vs. SHAP-FOIL.

\begin{table}[h]
{\scriptsize 

{\begin{tabular}{|c|c|c|c|c|c|c|c|c|c|}
\cline{1-10}
Dataset&credit-J&breast-w& ecoli& kidney &voting &autism &ionosphere&sonar&heart \\
\cline{1-10}
ALEPH & 30 & 20 & 13 & 10 & 7 & 11 & 13 & 10 & 13 \\
\cline{1-10}
SHAP-FOIL & 22 & 5 & 8 & 5 & 5 & 6 & 14 & 6 & 6\\
\cline{1-10}
\end{tabular}
}
}
\caption{Number of lines in logic programs induced by Aleph and SHAP-FOIL}
\label{tab:progsize}
\end{table}

It should be noted that the comprehensibility of the program generated by SHAP-FOIL is superior to those generated by various rule-extraction methods 

\section{Related Works}
\label{sec:related}

As noted earlier, all proposed Rule Extraction techniques either treat the model as black-box \cite{minerva,iter}, or are limited to certain type of kernels \cite{fung}, or are too complex to interpret \cite{svmprototype}. A survey of existing Rule Extraction techniques can be found in \cite{svmruleextraction}. 
In contrast, our SHAP-FOIL method produces a succinct logic program that has an elegant declarative as well as operational semantics resulting in better comprehensibility.
Most rule extraction methods adopt an ad hoc if-then-else notation for representing the rules they output.

Rule extraction from statistical Machine Learning models has been a long-standing goal of the community. The rule extraction algorithms from machine learning models are classified into two categories: 1) Pedagogical (i.e., learning symbolic rules from black-box classifiers without opening them) 2) Decompositional (i.e., to open the classifier and look into the internals). TREPAN \cite{trepan} is a successful pedagogical algorithm that learns decision trees from neural networks. Minerva \cite{minerva} and Iter \cite{iter} are pedagogical approaches to extract rules from SVM models. There is also a broader pedagogical approach to rule extraction where an SVM model is trained first, then the entire training data is re-labeled using the predictions of the SVM model, and finally a rule learning method (e.g., C4.5, ID3, CART etc.) is applied. It should be noted that this approach suffers from the very issue of local optima discussed in Figure \ref{fig:localoptima}.   

SVM+Prototypes \cite{svmprototype} is a decompositional rule extraction algorithm that makes use of KMeans clustering to extract rules from SVM classifiers by focusing on support vectors. The main drawback of this approach is that all extracted rules contain all possible input variables in its conditions, making the approach too complex to interpret for large input dimensions. Fung \cite{fung} is another decompositional SVM rule extraction technique to extract propositional rules which is limited to linear kernels. The advantage of our SHAP-FOIL algorithm is that it can handle all kernels and the induced hypotheses are expressed in terms of the original features.

\section{Conclusions and Future Work}
\label{sec:concl}
In this paper we presented an algorithm to open up the Support Vector Machine (SVM) models and induce logic programs by looking at the support vectors as the most influential points in making classification decision. We make use of SHAP \cite{shap} to find the most relevant features that would contribute to the support vector's prediction. We showed that finding data points ``similar" to influential support vectors allows same feature values to induce a hypothesis that correctly predicts internal data points. The resulting algorithm significantly outperforms state-of-the-art ILP system ALEPH in terms of the classification evaluation measures while it does not suffer from the local optima issue as explained earlier.

There are number of directions for future work: (i)  Equation \eqref{eq:similarity} can measure dissimilarity too. Dissimilarity is usually expressed as \textit{negation-as-failure} in ILP \cite{fold}. We plan to incorporate dissimilarity to support vectors of the opposite class as a way to reduce the number of induced clauses as well as increase the accuracy of each clause. In such a case, we will be learning answer set programs \cite{gelfondkahl}. 
Learning answer set programs has an added advantage of being able to distinguish between exceptions and noise \cite{fold,farhad-padl}.
(ii) In sparse regions of training data, specialization is usually stopped too early. If the algorithm can introduce synthesized data and have the SVM model label it, the SHAP-FOIL algorithm can continue specialization further and achieve better test accuracy. 
(iii) Our work thus far has focused on classification problems. We plan to generalize and extend our work to induce logic programs from regression models in the future. This will be achieved by discretization of continuous spaces. It should be noted that we have applied our technique to gradient-boosting models as well \cite{farhad-padl} with excellent results. 

\section*{Acknowledgment}
Authors acknowledge support from NSF grants IIS 1718945, IIS 1910131, IIP 1916206 and DARPA grant HR001119S0057-ARCOS-FP-036.

\bibliographystyle{ACM-Reference-Format}
\bibliography{mycitations}

\label{lastpage}
\end{document}